\documentclass[runningheads]{llncs}

\usepackage{eccv}

\usepackage{eccvabbrv}

\usepackage{graphicx}
\usepackage{booktabs}

\usepackage[accsupp]{axessibility}  % Improves PDF readability for those with disabilities.
\usepackage{graphicx}
\usepackage{amsmath}
\usepackage{amssymb}
\usepackage{mathtools}
\usepackage{booktabs}
\DeclareMathOperator{\SoftMax}{SoftMax}
\DeclareMathOperator{\Diag}{Diag}
\DeclareMathOperator{\tril}{tril}
\newcommand{\bdelta}[1]{\textcolor{ForestGreen!60!black}{\raisebox{0.25ex}{\scalebox{0.95}{(#1)}}}}
\newcommand{\rdelta}[1]{\textcolor{Red!60!black}{\raisebox{0.25ex}{\scalebox{0.95}{(#1)}}}}
\usepackage{amsmath,amssymb,mathtools}
\usepackage{bm}                 % bold math symbols
\usepackage{algorithm}
\usepackage{algpseudocode}      % algorithmicx
\usepackage{booktabs,multirow}  % (already used by your tables)
\usepackage{amsmath,amssymb,mathtools}
\usepackage{algorithm}
\usepackage{algpseudocode}
\usepackage{booktabs,multirow}

\usepackage{booktabs,multirow,siunitx,xcolor}
\usepackage{booktabs,siunitx,multirow}
\usepackage[table,svgnames]{xcolor}   % for RoyalBlue*
\usepackage{graphicx}                 % for \scalebox

\usepackage{hyperref}
\usepackage{xcolor}

\newcommand{\gd}[1]{\textcolor{green!60!black}{#1}}

\usepackage{orcidlink}

\begin{document}

% ---------------------------------------------------------------
% TODO REVIEW: Replace with your title
\title{See Fair, Speak Truth: Equitable Attention Improves Grounding and Reduces Hallucination in Vision-Language Alignment} 

% TODO REVIEW: If the paper title is too long for the running head, you can set
% an abbreviated paper title here. If not, comment out.
\titlerunning{Abbreviated paper title}

% TODO FINAL: Replace with your author list. 
% Include the authors' OCRID for the camera-ready version, if at all possible.
% -------------------------
% ECCV/LNCS: Authors
% -------------------------
\title{See Fair, Speak Truth: Equitable Attention Improves Grounding and Reduces Hallucination in Vision-Language Alignment}
\titlerunning{See Fair, Speak Truth}

% -------------------------
% Authors (ECCV/LNCS)
% -------------------------
\author{
Mohammad Anas Azeez\inst{1} \and
Ankan Deria\inst{1} \and
Zohaib Hasan Siddiqui\inst{2} \and
Adinath Madhavrao Dukre\inst{1} \and
Rafiq Ali\inst{3} \and
Sara Atito\inst{4} \and
Yutong Xie\inst{1} \and
Imran Razzak\inst{1,5}
}

\authorrunning{Azeez et al.}

\institute{
Mohamed bin Zayed University of Artificial Intelligence, Abu Dhabi, UAE \\
\and
King Fahd University of Petroleum and Minerals, Dhahran, Saudi Arabia \\
 \and
Macquarie University, Sydney, Australia \\
 \and
University of Surrey, Guildford, United Kingdom \\
 \and
University of New South Wales, Sydney, Australia
}
\maketitle

\begin{figure*}[ht!]
\centering
    \includegraphics[width=\textwidth]{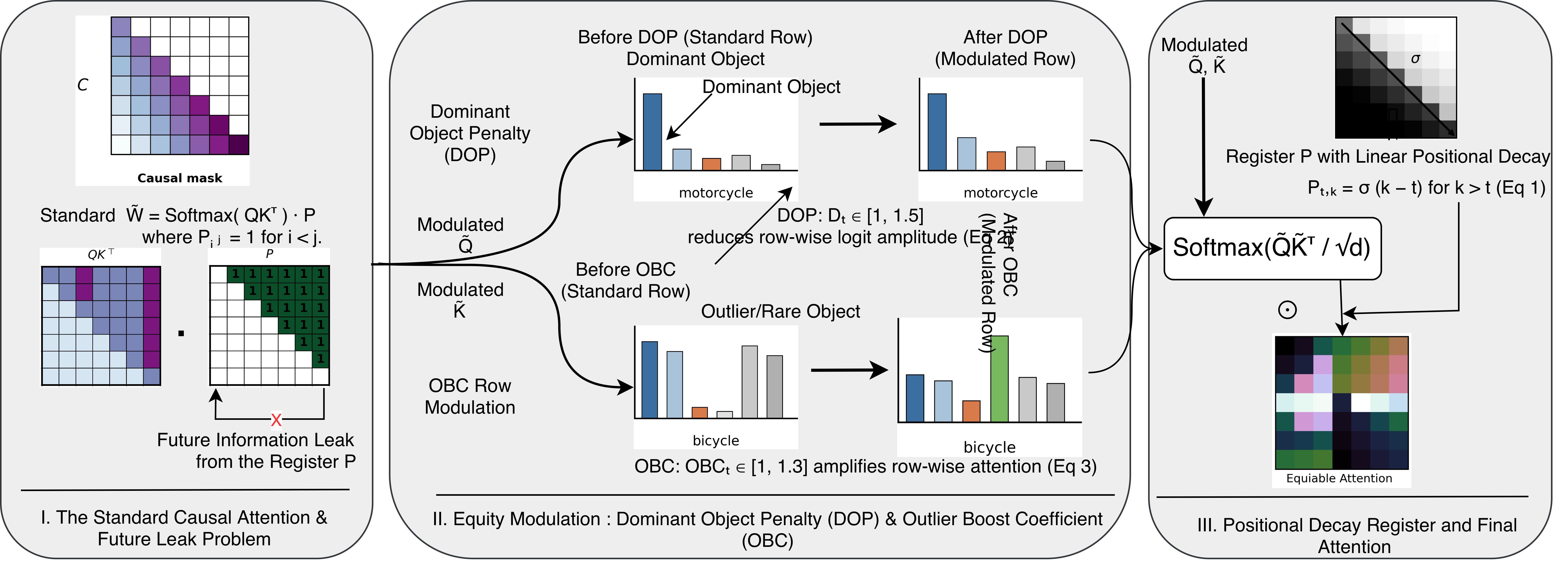}
    \caption{DOP–OBC Method Overview. (I) Standard causal masking in transformer attention. (II) Equity-aware modulation adjusts row-wise logit amplitudes: the Dominant Object Penalty (DOP) softly suppresses over-attended, visually dominant objects to free representational capacity, while the Outlier Boost Coefficient (OBC) amplifies rare yet confidently detected objects. (III) These object-aware signals are integrated with an upper-triangular positional decay register to form an equitable attention matrix at inference time, requiring no weight updates.}
    \label{fig:method}
    \vspace{-20 pt}
\end{figure*}

\begin{abstract}
Multimodal large language models (MLLMs) frequently hallucinate objects that are absent from the visual input, often because attention during decoding is disproportionately drawn to visually dominant or frequently occurring content. We observe that this inequity in attention allocation is a root cause of object hallucination: when rare, small, or contextually peripheral objects receive insufficient attention, the model fails to ground its generation in the full visual scene. We argue that every object in an image, regardless of its size, frequency or           visual salience, deserves equal representational opportunity during decoding. To this end, we propose \textbf{DOP-OBC}, a training-free and architecture-agnostic decoding strategy built on the principle of \emph{equitable attention}. Two complementary object-aware signals work in tandem: a \textbf{Dominant Object Penalty (DOP)} that softly suppresses attention over-concentration on visually dominant regions, and an \textbf{Outlier Boost Coefficient (OBC)} that amplifies attention toward rare yet confidently detected objects. These signals are injected as per-row logit modulations within the causal attention mask, requiring no weight updates and preserving autoregressive decoding properties. Extensive experiments across image and video MLLMs demonstrate consistent reductions in object hallucination on CHAIR and POPE benchmarks, alongside improvements in GPT-4o-assessed captioning quality across correctness, consistency, detail, context and temporal dimensions. DOP-OBC establishes that fairness in attention allocation is not merely a design principle but a practical and effective path toward more faithful multimodal generation.

\end{abstract}

\section{Introduction}

Multimodal Large Language Models (MLLMs) have achieved remarkable progress in vision--language understanding, demonstrating strong capabilities in image captioning, visual question answering, instruction following, and multimodal reasoning~\cite{liu2024llava15,liu2023llava,dai2023instructblip}. By combining powerful vision encoders with large language models, these systems generate fluent and contextually coherent descriptions across diverse domains. However, despite their impressive generative quality, a critical reliability gap persists: models frequently produce outputs that are only partially grounded in the visual evidence. Among the most prevalent and concerning failure modes is \emph{object hallucination}, where the model omits visually present objects or confidently describes objects that do not exist in the scene~\cite{rohrbach2018object,li2023pope}. The persistence of this issue across architectures and benchmarks suggests that the limitation is not purely representational, but fundamentally tied to how multimodal information is integrated and decoded.

\begin{figure}[t]
  \centering
  \includegraphics[width=1\columnwidth]{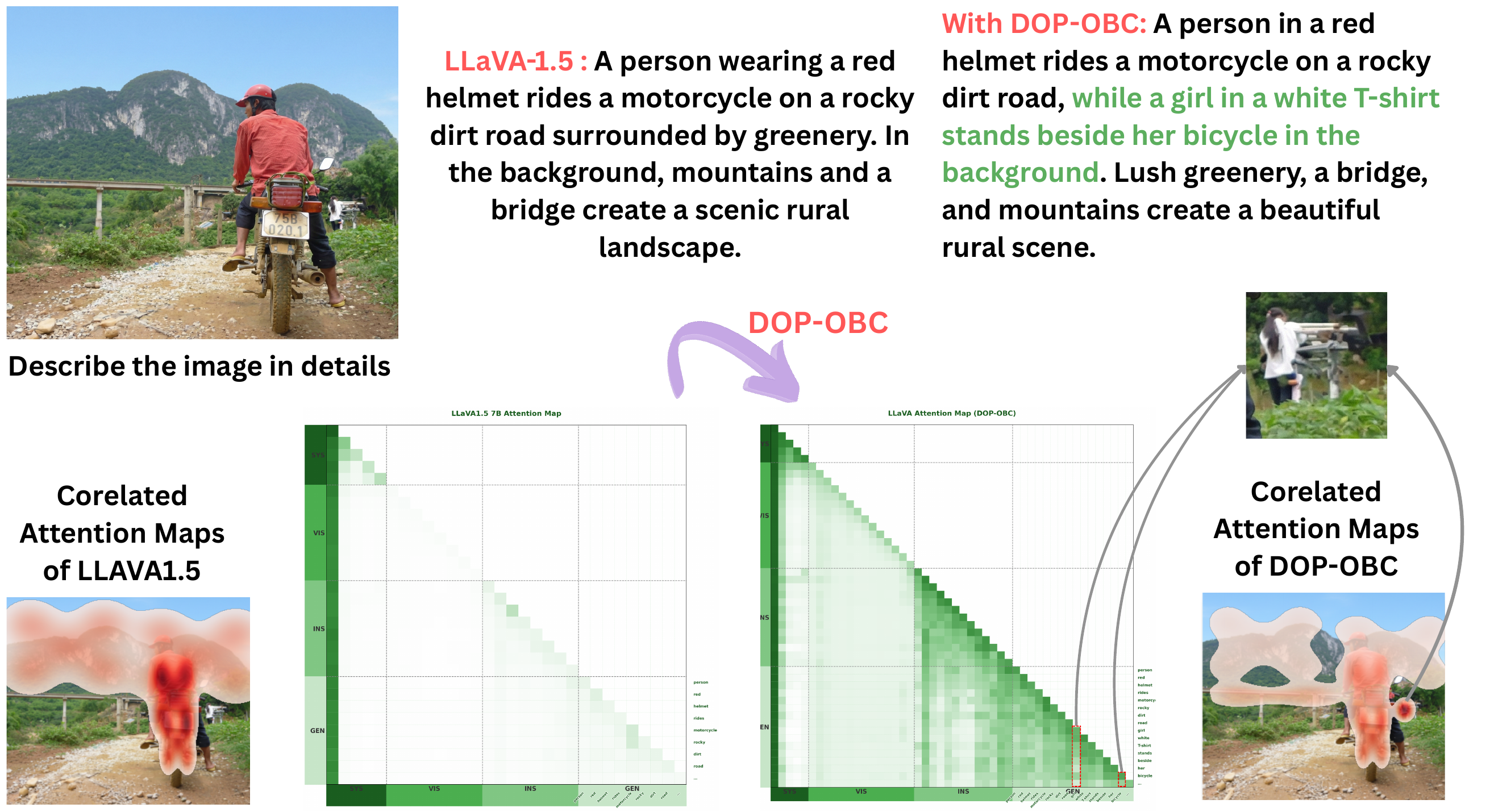}
  \caption{Example showing that DOP–OBC produces a more complete, grounded description than the LLaVA-1.5 baseline: the method adds correct background entities (highlighted in green in the figure text), and the accompanying attention-map comparison illustrates the reallocation of attention toward those previously neglected objects.} 
  \label{fig:firstpage}
  \vspace{-10 pt}
\end{figure}

Figure~\ref{fig:firstpage} illustrates this imbalance concretely. In the example, \textbf{LLaVA-1.5} correctly captures the dominant elements of the scene, namely the motorcycle and its rider, yet completely overlooks a clearly visible girl standing beside her bicycle. The correlated attention maps clarify the source of this failure. Although the visual encoder activates across multiple regions, the decoder progressively concentrates attention on the dominant foreground object. In the green attention weight plots, most visual tokens receive uniformly low weights, while a narrow band corresponding to the motorcycle region accumulates strong and persistent attention. As decoding advances, this concentration intensifies, leading to attention collapse and suppression of peripheral yet visually present objects. Importantly, this is not a perception failure, since the encoder representations already contain signals for the girl and the bicycle. Rather, the issue arises during decoding, where attention allocation becomes structurally biased toward large, salient, and frequently occurring objects due to token abundance and language priors. Simply redistributing attention equally does not resolve this structural imbalance, as equality assigns identical resources while equity compensates for disparities in scale and frequency.

To address this challenge, we propose \textbf{DOP--OBC}, a training-free and architecture-agnostic decoding strategy that explicitly rebalances attention at inference time. As summarized in Figure~\ref{fig:method}, our approach injects two complementary object-aware controls directly into the attention computation. The \textbf{Dominant Object Penalty (DOP)} suppresses over-attended objects by reducing their row-wise logit amplitude, while the \textbf{Outlier Boost Coefficient (OBC)} amplifies rare yet confidently detected objects through calibrated scaling. These modulations are integrated within an upper-triangular positional decay register that preserves autoregressive causality while stabilizing long-range grounding. In Figure~\ref{fig:firstpage}, this mechanism suppresses excessive focus on the motorcycle and elevates the underrepresented girl and bicycle. The resulting attention maps show a more balanced and sustained green weight distribution across decoding steps, enabling faithful grounding of all relevant objects. Importantly, this reallocation operates entirely at inference time and requires no retraining, auxiliary models, or architectural modification.

The main contributions of this paper are as follows:

\begin{itemize}

    \item We identify a systematic attention imbalance in existing MLLMs, where visually dominant objects consume disproportionate decoding capacity, suppressing small, rare, or peripheral objects. We show that this inequity is a key driver of object hallucination and remains unaddressed by prior methods.

    \item We propose DOP-OBC, a training-free and architecture-agnostic decoding strategy that corrects this imbalance through two complementary plug-in signals: a Dominant Object Penalty that attenuates over-attended objects and an Outlier Boost Coefficient that amplifies rare yet confidently detected ones. Both operate as per-row modulations within an upper-triangular attention register without modifying model weights.

    \item We validate DOP-OBC across multiple image and video MLLMs, reducing hallucination on CHAIR and POPE and improving all five GPT-4o-evaluated video captioning dimensions. These results show that equitable attention allocation is an effective inference-time approach for faithful multimodal generation.

\end{itemize}

\section{Related Work}
\label{sec:related}

\subsection{Vision-language models}
% Recent progress in multimodal large language models (MLLMs) has markedly advanced vision--language integration, strengthening multimodal reasoning, instruction following, and interactive capabilities \cite{liu2023llava,liu2024llava15,li2023blip}. LLaVA augments LLaMA with a visual encoder, enabling image-conditioned instruction tuning and open-ended generation \cite{liu2023llava}. Flamingo introduces a few-shot architecture that interleaves visual and textual tokens for contextual multimodal reasoning \cite{alayrac2022flamingo}. InstructBLIP builds upon BLIP-2 through instruction tuning, further improving image-grounded dialogue and general multimodal understanding \cite{dai2023instructblip}. Extending these advances to video, Video-ChatGPT adapts GPT-style models for conversational video analysis, while Video-LLaVA incorporates temporal modeling into the LLaVA framework to support coherent video dialogue and comprehension \cite{lin2024videollava}. 

Recent progress in multimodal large language models (MLLMs) has substantially advanced vision--language integration, improving multimodal reasoning, instruction following, and conversational capabilities~\cite{liu2023llava,liu2024llava15,dai2023instructblip}. LLaVA aligns a pretrained LLM with a visual encoder through instruction tuning, enabling open-ended image-conditioned generation~\cite{liu2023llava}. Flamingo introduces an interleaved visual-text token architecture that supports few-shot multimodal reasoning~\cite{alayrac2022flamingo}. InstructBLIP extends BLIP-2 with instruction tuning to enhance image-grounded dialogue and general multimodal understanding~\cite{dai2023instructblip}. For video understanding, Video-LLaVA and related frameworks incorporate temporal modeling into LLaVA-style architectures to enable coherent video dialogue and reasoning~\cite{lin2024videollava}. 
Despite these advances, object hallucination and grounding failures remain persistent challenges. Rohrbach et al.~\cite{rohrbach2018object} first systematically analyzed object hallucination in image captioning, showing that language priors often override visual evidence. POPE~\cite{li2023pope} further demonstrates that modern MLLMs frequently generate objects unsupported by the input image, even when confidence is high. More recently, FarSight~\cite{tang2025farsight} identifies decoding-time attention collapse and positional drift as key contributors to hallucination in LLaVA-style models. These findings indicate that hallucination is not solely a representation issue but is closely tied to attention allocation during decoding, motivating inference-time attention rebalancing strategies such as ours.

\subsection{Contrastive decoding and attention modification}
A major line of inference-time mitigation for object hallucination in multimodal models is contrastive decoding, which suppresses spurious predictions by contrasting distributions from perturbed and original inputs. Visual Contrastive Decoding (VCD) contrasts logits from original and distorted visual inputs to reduce over-reliance on language priors and improve visual grounding without retraining \cite{leng2024mitigating,yin2025clearsight}. Extensions of this paradigm include Attention-Steerable Contrastive Decoding (ASCD), which directly steers attention scores during decoding to reduce hallucination by redistributing cross-modal attention \cite{wang2025ascd}, and multi-frequency contrastive decoding (MFCD), which removes hallucination components by targeting frequency content in the output distribution \cite{liu2025multi}. Inter-layer attention contrastive methods (e.g., iTaD) select intermediate representations with strong image token focus to contrast and highlight genuine visual signals for more faithful outputs \cite{xu2025mitigating}. Related work also proposes enhancing visual signals within modality fusion layers to counteract language bias in contrastive setups (e.g., Visual Amplification Fusion) \cite{yin2025clearsight}. These methods operate at inference time and focus on modifying attention or output distributions, complementing our approach of directly rebalancing object-level attention via dominance suppression and rarity-aware boosts.

\subsection{Externally guided decoding}
Another line of work reduces hallucination by incorporating external guidance at inference time. CLIP-guided decoding uses pretrained contrastive vision–language embeddings to steer generation toward visually grounded token predictions, improving alignment between images and captions \cite{deng2024seeing}. Summary-guided decoding modifies the decoding process by using concise summaries to focus the model on image-relevant content, which can reduce language prior dominance \cite{min2025mitigating}. Interaction guidance sampling explicitly guides the model to leverage multimodal interaction information during generation, reducing spurious predictions without additional training \cite{dong2025inter}. Other approaches train auxiliary estimators, such as value models, to score candidate continuations and bias search toward low-hallucination outputs \cite{wang2025scaling,deria2025dual}. These externally guided strategies differ from our approach by relying on additional models or external signals; in contrast, we derive guidance solely from internal object proposals and attention statistics to provide targeted, object-level control at inference time.

\section{Methodology}
\label{sec:method}

We propose a fully inference-time decoding strategy that improves attention allocation without any finetuning or architectural modification. The method operates directly within the attention computation and consists of three components. 
First, attention registers absorb surplus attention assigned to future positions, steering decoding toward contextually informative tokens (Sec.~\ref{sec:registers}). Second, a progressively diminishing masking rate introduces absolute positional focus while preserving autoregressive causality (Sec.~\ref{sec:posenc}). Third, to ensure equitable allocation across long-tailed objects, we introduce two lightweight object-aware controls: the Dominant Object Penalty (DOP) and the Outlier Boost Coefficient (OBC), which rebalance attention between dominant and under-represented objects (Sec.~\ref{sec:equisight}). 
% PyTorch-style pseudo-code is provided in Alg.~\ref{alg:equisight}.

\subsection{Upper-Triangular Attention Registers}
\label{sec:registers}

To mitigate attention collapse, we introduce an upper-triangular attention register that absorbs surplus mass assigned to future positions. Let $\bm{\omega} \in \mathbb{R}^{n \times n}$ denote the query–key score matrix and let $\mathbf{C} = \tril(\mathbf{1})$ be the standard lower-triangular causal mask. We modify the logits as
\begin{equation}
\label{eq:reg-w}
\mathbf{W} \;=\; \bm{\omega}\mathbf{C} \;+\; \mathbf{P},
\qquad
\mathbf{P}_{i,j} =
\begin{cases}
-(j-i)\sigma, & j > i, \\
0, & \text{otherwise},
\end{cases}
\end{equation}
where $\mathbf{P} \in \mathbb{R}^{n \times n}$ is an upper-triangular matrix with linear decay slope $\sigma > 0$. The term $\mathbf{P}$ provides a controlled reservoir for excess logits directed toward future tokens, preventing unintended redistribution over valid positions after normalization.

The masked attention is then computed as
\begin{equation}
\label{eq:final-attn}
\widetilde{\mathbf{W}} \;=\; \SoftMax(\mathbf{W}) \mathbf{C}.
\end{equation}

\noindent\textbf{Remark.} The mask $\mathbf{C}$ in Eq.~\eqref{eq:reg-w} constrains logits before normalization, whereas the mask in Eq.~\eqref{eq:final-attn} removes residual future mass after normalization, ensuring strictly causal decoding.

\begin{algorithm}[t]
\caption{DOP--OBC Decoding (Single Decoder Layer)}
\label{alg:equisight}
\scriptsize
\begin{algorithmic}[1]
\Require Hidden states $\mathbf{x}$; causal mask $\mathbf{C}$ (lower-triangular)
\Require Object set $\mathcal{O}$ with features $\{\text{size}, \text{len}, \text{attn}, \text{conf}\}$
\Require Row-to-object alignment map $\pi(i)$
\Require Hyperparameters $\alpha_0, \sigma_0, \mathbf{w}=[w_1,w_2,w_3], \lambda, \gamma, \tau_p, \tau_r, r_{\max}$
\Ensure Updated decoder output $\mathbf{y}$

\Statex
\State \textbf{1. QKV Projection and Raw Attention Scores}
\State $(\mathbf{Q}, \mathbf{K}, \mathbf{V}) \gets \texttt{qkv\_proj}(\mathbf{x})$
\State $\mathbf{S} \gets (\mathbf{Q}\mathbf{K}^\top)/\sqrt{d_k}$

\Statex
\State \textbf{2. Dominant Object Penalty (DOP)}
\State Normalize object statistics:
\[
\hat{\mathbf{s}}, \hat{\mathbf{L}}, \hat{\mathbf{A}} 
\gets \texttt{minmax}(\text{size}),\;
      \texttt{minmax}(\text{len}),\;
      \texttt{minmax}(\text{attn})
\]
\State $\mathbf{D} \gets w_1\hat{\mathbf{s}} + w_2\hat{\mathbf{L}} + w_3\hat{\mathbf{A}}$
\State $\mathbf{w}_{\text{pen}} \gets \exp(-\lambda \mathbf{D})$

\Statex
\State \textbf{3. Outlier Boost Coefficient (OBC)}
\State $(\boldsymbol{\mu}, \mathbf{\Sigma}) \gets \texttt{ema\_gaussian}(\text{feat})$
\State $\mathbf{r} \gets \texttt{clamp}\!\left(
\frac{\texttt{mahalanobis}(\text{feat}, \boldsymbol{\mu}, \mathbf{\Sigma})}{\tau_r},\;
0,\; r_{\max}
\right)$
\State $\mathbf{b} \gets \gamma \mathbf{r}\, \mathbb{1}[\text{conf} \ge \tau_p]$

\Statex
\State \textbf{4. Row-wise Modulation and Register Construction}
\State Initialize $\boldsymbol{\alpha} \gets \alpha_0 \mathbf{1}$,\quad
$\boldsymbol{\sigma} \gets \sigma_0 \mathbf{1}$
\For{$i = 1$ to $n$}
    \State $o \gets \pi(i)$
    \If{$o \ge 0$}
        \State $\alpha_i \gets \alpha_i \cdot \mathbf{w}_{\text{pen}}[o] \cdot (1 + b_o)$
        \State $\sigma_i \gets \sigma_i / (1 + b_o)$
    \EndIf
\EndFor
\State $\mathbf{P} \gets \texttt{build\_register}(n, \boldsymbol{\sigma})$

\Statex
\State \textbf{5. Attention Composition and Output}
\State $\mathbf{S} \gets (\text{diag}(\boldsymbol{\alpha})\,\mathbf{S}) \odot \mathbf{C} + \mathbf{P}$
\State $\mathbf{A} \gets \text{SoftMax}(\mathbf{S}) \odot \mathbf{C}$
\State $\mathbf{y} \gets \texttt{wo}(\mathbf{A}\mathbf{V})$
\State \Return $\mathbf{y}$

\end{algorithmic}
\end{algorithm}
%-------------------------------------------------------------------------

\subsection{Positional Awareness Encoding}
\label{sec:posenc}
The linear decay in $\mathbf{P}$ produces a \emph{progressively diminishing} allocation to future tokens across rows: as the row index increases, the normalized mass distributed over valid (past) tokens grows monotonically. This yields absolute positional awareness without altering model parameters. Practically, later tokens aggregate more historical context while future positions are softly discouraged, stabilizing long-range generation.

\subsection{Equity-Aware Object Controls (DOP \& OBC)}
\label{sec:equisight}

To reallocate decoding capacity toward small and rare-but-valid objects, we introduce two object-aware signals that act through the same mechanisms as the attention register: a row-wise amplitude $\alpha_i$ that scales pre-SoftMax logits and a row-wise decay $\sigma_i$ that controls the register slope. Each attention row $i$ is associated with an object proposal $o(i)$ via the same ROI/patch assignment used during routing; rows without an aligned object retain default parameters.

\paragraph{Dominant Object Penalty (DOP).}
For each object $o$, we compute a normalized dominance score
\begin{equation}
D(o) = w_1 \hat{s}(o) + w_2 \hat{L}(o) + w_3 \hat{A}(o),
\end{equation}
where $w_i \ge 0$ and $\sum_i w_i = 1$. Here, $\hat{s}(o)$ denotes normalized object size, $\hat{L}(o)$ its temporal persistence, and $\hat{A}(o)$ the attention share already received, each min–max normalized within the current window. A soft penalty
\begin{equation}
w_{\text{pen}}(o) = \exp\!\big(-\lambda D(o)\big), \quad \lambda \ge 0,
\end{equation}
attenuates attention allocated to dominant objects, freeing representational capacity for less prominent instances.

\paragraph{Outlier Boost Coefficient (OBC).}
To promote rare yet reliable objects, we estimate a rarity score $r(o)$ using a clipped, temperature-scaled distance between object features and an exponential moving average scene distribution. The boost is gated by posterior confidence $p(o)$:
\begin{equation}
b(o) = \gamma\, r(o)\, \mathbf{1}\!\left[p(o) \ge \tau_p\right],
\end{equation}
so that only rare and confident objects receive amplification.

\paragraph{Integration into attention.}
For each row $i$ linked to object $o(i)$, the controls are applied as
\begin{equation}
\alpha_i \leftarrow \alpha_0\, w_{\text{pen}}\!\big(o(i)\big)\, \big(1 + b(o(i))\big),
\qquad
\sigma_i \leftarrow \frac{\sigma_0}{1 + b(o(i))}.
\end{equation}
We replace $\bm{\omega}$ with $\Diag(\bm{\alpha})\,\bm{\omega}$ in Eq.~\eqref{eq:reg-w} and construct the register $\mathbf{P}$ using per-row slopes $\{\sigma_i\}$, while Eq.~\eqref{eq:final-attn} remains unchanged. 

In effect, DOP dampens attention to already dominant objects, and OBC amplifies and prolongs attention to rare but credible ones, ensuring that long-tail objects are less likely to be suppressed during decoding.

\subsection{Decoder Layer Composition}

Each decoder layer applies the following sequence: (i) QKV projection; (ii) row-wise logit scaling via $\Diag(\bm{\alpha})$; (iii) register construction with per-row decay; (iv) masked SoftMax normalization; and (v) value aggregation and projection, as summarized in Alg.~\ref{alg:equisight}. An optional pre-routing step rescales proposal scores using the same DOP and OBC signals to align selection with attention reallocation.

\subsection{Practical Notes and Defaults}
Per-window normalization uses a small $\varepsilon$ for stability. EMA density is vectorizable; $k$NN can be substituted. Default knobs: $w{=}\langle0.5,0.25,0.25\rangle$, $\lambda{=}1.0$, $\tau_p{=}0.5$, $\gamma{=}0.3$, $r_{\max}{=}2.0$. Safeguards: clip $b$, floor $\alpha_i$, and bound $\sigma_i\!\in\![\sigma_{\min},\sigma_0]$. The added cost is $O(|\mathcal{P}_t|+n)$; the attention complexity is unchanged.

\section{Experiments}
\label{sec:exp}

\begin{figure}[t!]
    \centering
    \includegraphics[width=\columnwidth]{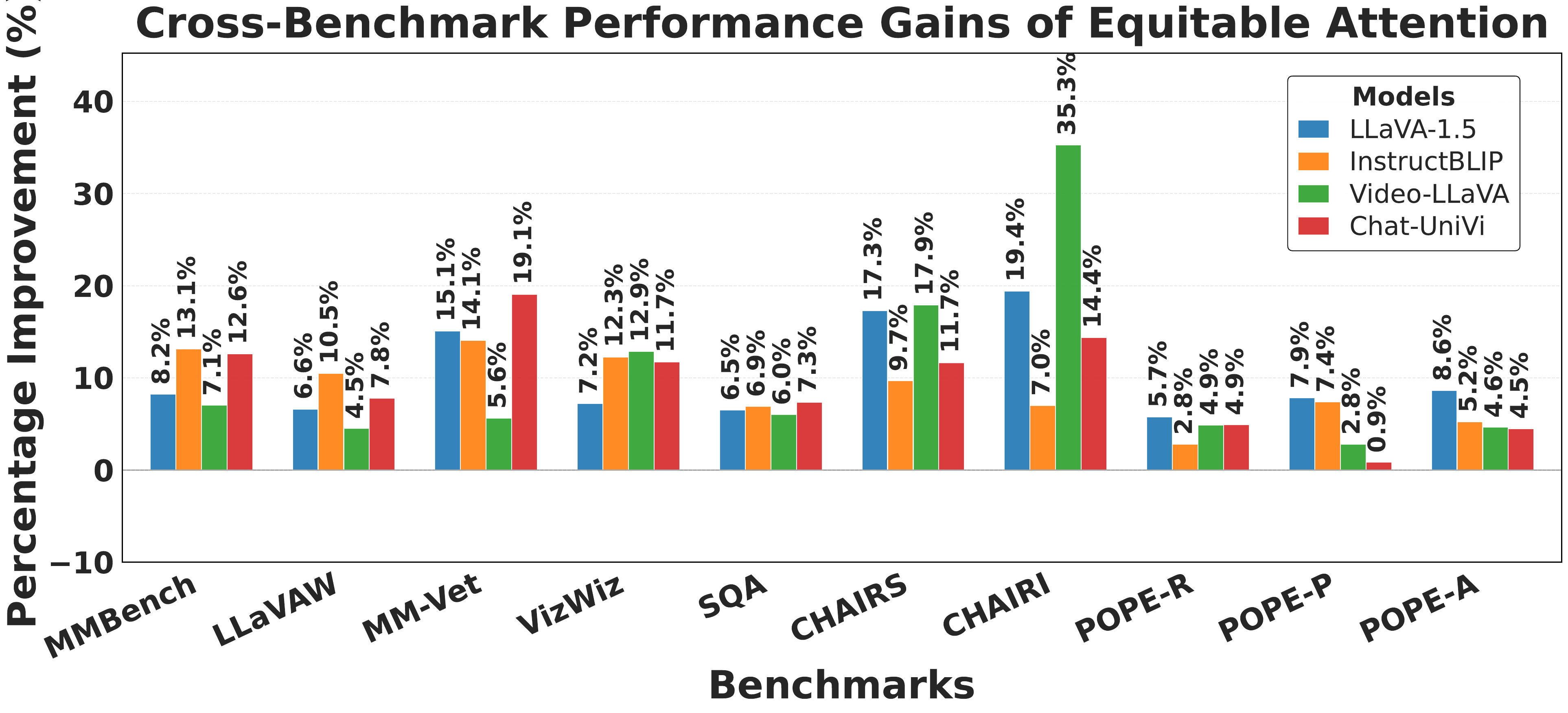}
    % \caption{\textbf{Cross-Benchmark Symphony of Gains.} DOP-OBC orchestrates consistent performance uplifts across MMBench, LLaVAW, MM-Vet, VizWiz, SQA, and dramatic reductions in CHAIRS/CHAIRI, illustrating the harmony between fairness and precision in multimodal attention.}
    \caption{Cross-benchmark percentage improvement of DOP–OBC over base models across general multimodal benchmarks and hallucination metrics, showing consistent gains in task performance and reductions in CHAIR/POPE errors.}
    \label{Figure 3}
    \vspace{-10 pt}
\end{figure}

We evaluate our equity-aware decoding framework on both \emph{image} and \emph{video} benchmarks, following established protocols for hallucination mitigation and instruction-tuned MLLMs \cite{liu2023llava,rohrbach2018object,li2023pope,tang2025farsight}. Unless otherwise specified, we adopt official dataset splits and prompts, and we retain each base model’s default decoding settings, including temperature. Our method operates as a training-free, drop-in plug-in applied solely at inference time.

\subsection{Implementation Details}
We keep all model weights frozen and apply our plug-ins at decode time. Object proposals/tracks and confidences come from each model’s standard vision stack; our row-wise amplitude/decay are computed on-the-fly and vectorized. Default knobs ($\lambda,\gamma,\tau_p$ etc.) are as in Sec.~\ref{sec:method}. No extra training data or finetuning is used.

\subsection{Benchmarks and Metrics}

\textbf{Image Benchmarks.}
We evaluate general multimodal reasoning and hallucination behavior on widely adopted image benchmarks, with quantitative results reported in Table~\ref{tab:image-bench} and cross-benchmark improvements illustrated in Figure~\ref{Figure 3}. The evaluation suite includes MMBench~\cite{liu2023mmbench}, LLaVA-Bench-in-the-Wild (LLaVA-W)~\cite{liu2023llava}, MM-Vet~\cite{yu2023mm}, VizWiz~\cite{gurari2018vizwiz}, ScienceQA (SQA)~\cite{lu2022scienceqa}, and the hallucination-focused CHAIR and POPE datasets~\cite{rohrbach2018object,li2023pope}. For CHAIR, we report $\text{CHAIRS}\downarrow$ and $\text{CHAIRI}\downarrow$, where lower values indicate fewer hallucinations. For POPE, we report Recall, Precision, and Accuracy (R/P/A) $\uparrow$, where higher values reflect stronger visual grounding and reduced object-level hallucination.

% in your Experiments section (after the CHAIR/POPE paragraph)
\begin{figure}[t]
  \centering
  \includegraphics[width=0.8\columnwidth]{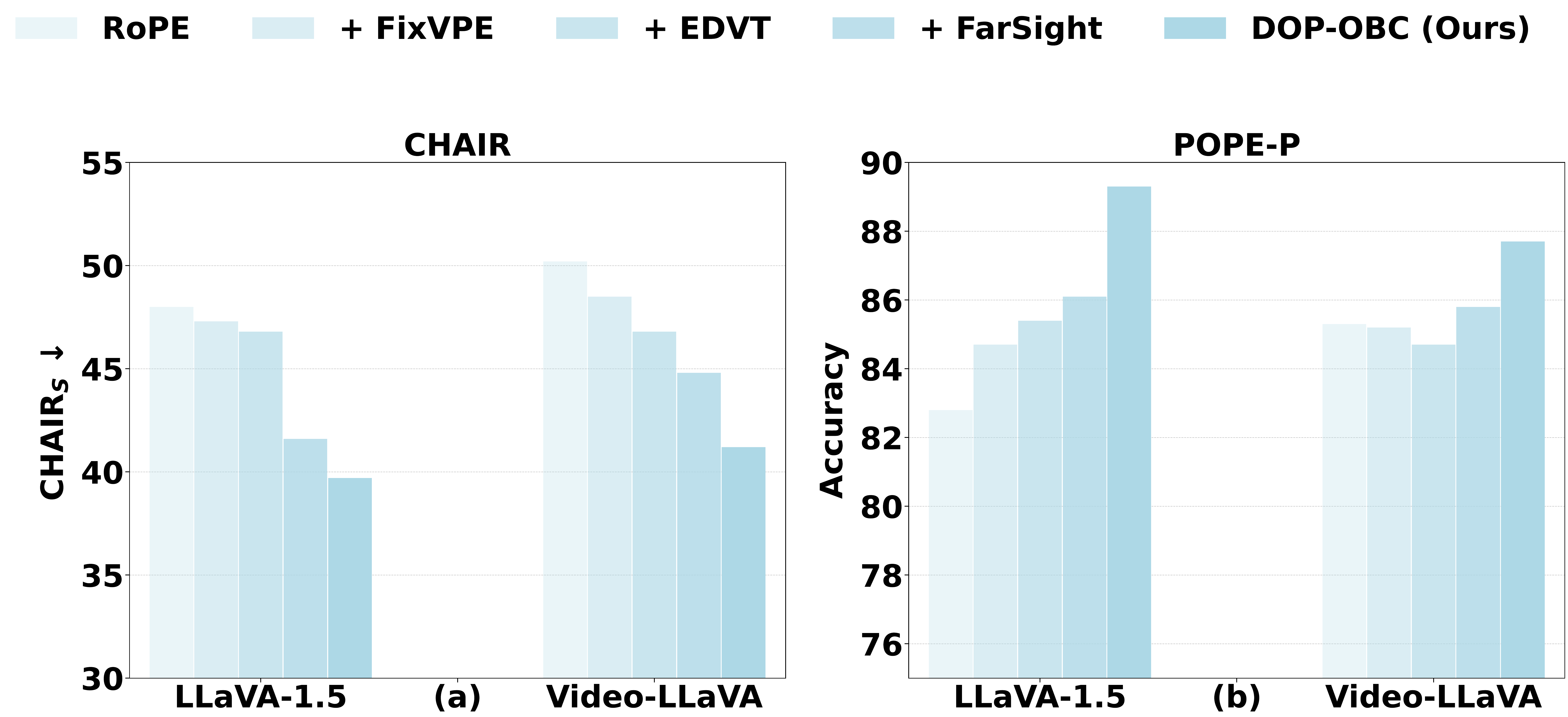}
  \caption{\textbf{CHAIR and POPE-P comparisons on LLaVA-1.5 and Video-LLaVA.}
  We report CHAIR$_S$ (lower is better) and POPE-P (higher is better) under RoPE, FixVPE, EDVT, FarSight, and \textbf{DOP-OBC(Ours)}.
  The rightmost bars (\textbf{DOP-OBC(Ours)}) achieve the largest improvements across both backbones.}
  \label{fig:chair-pope}
  \vspace{-10pt}
\end{figure}

\begin{table}[t]
\centering
\caption{Comparison of decoding methods on CHAIR and 
POPE benchmarks. $\downarrow$ indicates lower is better; 
$\uparrow$ indicates higher is better. $\Delta$ denotes 
improvement of DOP--OBC over the respective LLAVA-1.5 baseline.}
\label{tab:positional}
\setlength{\tabcolsep}{8pt}
\renewcommand{\arraystretch}{1.2}
\resizebox{0.8\textwidth}{!}{
\begin{tabular}{ll cccc}
\toprule
\textbf{Base Model} 
    & \textbf{Method} 
    & \textbf{CHAIR}$_S$ $\downarrow$ 
    & \textbf{CHAIR}$_I$ $\downarrow$ 
    & \textbf{POPE-R} $\uparrow$ 
    & \textbf{POPE-P} $\uparrow$ \\
\midrule

\multirow{5}{*}{\shortstack[l]{LLaVA-1.5 \\ 
    \cite{liu2024llava15}}}
    & LLaVA-1.5 (RoPE)      & 48.0 & 13.9 & 87.0 & 82.8 \\
    & FixVPE                 & 47.3 & 13.4 & 87.5 & 84.7 \\
    & EDVT                   & 46.8 & 14.5 & 87.8 & 85.4 \\
    & FarSight               & 41.6 & 13.2 & 90.5 & 86.1 \\
    \cmidrule(lr){2-6}
    & DOP--OBC (Ours) 
        & \textbf{39.7}{\scriptsize\gd{$_{(\Delta{-8.3})}$}}
        & \textbf{11.2}{\scriptsize\gd{$_{(\Delta{-2.7})}$}}
        & \textbf{92.0}{\scriptsize\gd{$_{(\Delta{+5.0})}$}}
        & \textbf{89.3}{\scriptsize\gd{$_{(\Delta{+6.5})}$}}\\

\midrule

\multirow{5}{*}{\shortstack[l]{Video-LLaVA \\ 
    \cite{lin2024videollava}}}
    & Video-LLaVA (RoPE)    & 50.2 & 15.6 & 81.6 & 85.3 \\
    & FixVPE                & 48.5 & 14.9 & 81.9 & 85.2 \\
    & EDVT                  & 46.8 & 13.7 & 82.5 & 84.7 \\
    & FarSight              & 44.8 & 12.9 & 83.2 & 85.8 \\
    \cmidrule(lr){2-6}
    & DOP--OBC (Ours) 
        & \textbf{41.2}{\scriptsize\gd{$_{(\Delta{-9.0})}$}}
        & \textbf{10.1}{\scriptsize\gd{$_{(\Delta{-5.5})}$}}
        & \textbf{85.6}{\scriptsize\gd{$_{(\Delta{+4.0})}$}}
        & \textbf{87.7}{\scriptsize\gd{$_{(\Delta{+2.4})}$}}\\

\bottomrule
\end{tabular}}
\end{table}

\begin{table*}[t]
\centering
\caption{Video benchmarks. For Video-Based Text Generation we report GPT-4o–assessed metrics:
Correctness (Cr.), Consistency (Cs.), Detail (De.), Context (Ct.), Temporal (Te.). Higher is better.}
\label{tab:video-bench}
\begingroup
\setlength{\tabcolsep}{4.5pt}
\renewcommand{\arraystretch}{0.95}
\small
\resizebox{\textwidth}{!}{
\begin{tabular}{l|l|ll|ll|lllll}
\toprule
\textbf{Base Model} & \textbf{Method}
& \multicolumn{2}{c|}{\textbf{MSVD-QA}}
& \multicolumn{2}{c|}{\textbf{ActivityNet-QA}}
& \multicolumn{5}{c}{\textbf{Video-Based Text Generation}} \\
\cmidrule(lr){3-4}\cmidrule(lr){5-6}\cmidrule(lr){7-11}
&
& {Accuracy $\uparrow$} & {Score $\uparrow$}
& {Accuracy $\uparrow$} & {Score $\uparrow$}
& {Cr. $\uparrow$} & {Cs. $\uparrow$} & {De. $\uparrow$} & {Ct. $\uparrow$} & {Te. $\uparrow$} \\
\midrule
\multirow{2}{*}{Char-UniVi \cite{chen2023chatunivi}}
& Base & 64.6 & 3.6 & 43.1 & 3.2 & 2.84 & 2.93 & 2.55 & 3.16 & 2.43 \\
& DOP-OBC & \textbf{69.2}\textsuperscript{\bdelta{+4.6}} & \textbf{4.0}\textsuperscript{\bdelta{+0.4}} & 43.0\textsuperscript{\rdelta{-0.1}}  & \textbf{3.6}\textsuperscript{\bdelta{+0.4}} & \textbf{2.90}\textsuperscript{\bdelta{+0.06}} & \textbf{3.00}\textsuperscript{\bdelta{+0.07}} & \textbf{2.62}\textsuperscript{\bdelta{+0.07}} & \textbf{3.30}\textsuperscript{\bdelta{+0.14}} & \textbf{2.56}\textsuperscript{\bdelta{+0.13}} \\
\midrule
\multirow{2}{*}{Video-LLaVA \cite{lin2024videollava}}
& Base & 64.8 & 3.7 & 41.5 & 3.3 & 2.32 & 2.34 & 2.65 & 2.75 & 2.09 \\
& DOP-OBC & \textbf{72.2}\textsuperscript{\bdelta{+7.4}} & \textbf{4.1}\textsuperscript{\bdelta{+0.4}} & \textbf{44.2}\textsuperscript{\bdelta{+2.7}} & \textbf{3.8}\textsuperscript{\bdelta{+0.5}} & \textbf{2.97}\textsuperscript{\bdelta{+0.65}} & \textbf{2.65}\textsuperscript{\bdelta{+0.31}} & \textbf{3.12}\textsuperscript{\bdelta{+0.47}} & \textbf{3.21}\textsuperscript{\bdelta{+0.46}} & \textbf{2.47}\textsuperscript{\bdelta{+0.38}} \\
\midrule
\multirow{2}{*}{VILA \cite{zhang2023vila}}
& Base & 72.6 & 4.0 & 50.2 & 3.3 & 3.14 & 3.40 & 2.71 & 3.43 & 2.58 \\
& DOP-OBC & \textbf{76.9}\textsuperscript{\bdelta{+4.3}} & \textbf{4.7}\textsuperscript{\bdelta{+0.7}} & \textbf{53.2}\textsuperscript{\bdelta{+3.0}} & \textbf{3.9}\textsuperscript{\bdelta{+0.6}} & \textbf{3.23}\textsuperscript{\bdelta{+0.09}} & \textbf{3.76}\textsuperscript{\bdelta{+0.36}} & \textbf{2.83}\textsuperscript{\bdelta{+0.12}} & \textbf{3.77}\textsuperscript{\bdelta{+0.34}} & \textbf{2.78}\textsuperscript{\bdelta{+0.20}} \\
\midrule
\multirow{2}{*}{Video-LLAMA2 \cite{ma2024videollama2}}
& Base & 70.9 & 3.8 & 49.9 & 3.3 & 3.13 & 3.23 & 2.70 & 3.42 & 2.45 \\
& DOP-OBC & \textbf{74.3}\textsuperscript{\bdelta{+3.4}} & \textbf{4.5}\textsuperscript{\bdelta{+0.7}} & \textbf{52.2}\textsuperscript{\bdelta{+2.3}} & \textbf{4.0}\textsuperscript{\bdelta{+0.7}} & \textbf{3.35}\textsuperscript{\bdelta{+0.22}} & \textbf{3.42}\textsuperscript{\bdelta{+0.19}} & \textbf{3.37}\textsuperscript{\bdelta{+0.67}} & \textbf{3.89}\textsuperscript{\bdelta{+0.47}} & \textbf{2.56}\textsuperscript{\bdelta{+0.11}} \\
\bottomrule
\end{tabular}}
\endgroup
\end{table*}

\begin{table*}[t]
\small
\centering
\caption{Image benchmarks. Lower is better for \textbf{CHAIRS}/\textbf{CHAIRI}; higher is better for all others.}
\label{tab:image-bench}
\resizebox{\textwidth}{!}{
\renewcommand{\arraystretch}{1.1}
\begin{tabular}{l|l|
l l l l l
l l l l l}
\toprule
\textbf{Base Model} & \textbf{Method}
& {\textbf{MMBench} $\uparrow$}
& {\textbf{LLaVAW} $\uparrow$}
& {\textbf{MM\textnormal{-}Vet} $\uparrow$}
& {\textbf{VizWiz} $\uparrow$}
& {\textbf{SQA} $\uparrow$}
& {\textbf{CHAIRS} $\downarrow$}
& {\textbf{CHAIRI} $\downarrow$}
& {\textbf{POPE\textnormal{-}R} $\uparrow$}
& {\textbf{POPE\textnormal{-}P} $\uparrow$}
& {\textbf{POPE\textnormal{-}A} $\uparrow$} \\
\midrule
\multirow{6}{*}{LLaVA-1.5 \cite{liu2024llava15}}
& Base    & 64.3 & 72.5 & 30.5 & 48.5 & 64.5 & 48.0 & 13.9 & 87.0 & 82.8 & 76.6 \\
& ICD     & 63.1 & 69.7 & 30.4 & 46.9 & 62.8 & 47.7 & 13.6 & 87.9 & 84.0 & 80.2 \\
& VCD     & 63.9 & 70.9 & 29.5 & 43.4 & 63.3 & 46.8 & 13.2 & 87.0 & 83.5 & 78.1 \\
& OPERA   & 64.4 & 72.0 & 31.4 & 50.0 & 64.9 & 45.2 & 12.7 & 88.8 & 82.8 & 79.2 \\
& FarSight& 66.0 & 74.7 & 32.5 & 50.8 & 67.4 & 41.6 & 13.2 & 90.5 & 86.1 & 80.4 \\
& DOP-OBC   & \textbf{69.6}\textsuperscript{\bdelta{+5.3}} & \textbf{77.3}\textsuperscript{\bdelta{+4.8}} & \textbf{35.1}\textsuperscript{\bdelta{+4.6}} & \textbf{52.0}\textsuperscript{\bdelta{+3.5}} & \textbf{68.7}\textsuperscript{\bdelta{+4.2}} & \textbf{39.7}\textsuperscript{\bdelta{+8.3}} & \textbf{11.2}\textsuperscript{\bdelta{+2.7}} & \textbf{92.0}\textsuperscript{\bdelta{+5.0}} & \textbf{89.3}\textsuperscript{\bdelta{+6.5}} & \textbf{83.2}\textsuperscript{\bdelta{+6.6}} \\
\midrule
\multirow{2}{*}{Video-LLaVA \cite{lin2024videollava}}
& Base    & 60.9 & 73.1 & 32.0 & 48.1 & 64.6 & 50.2 & 15.6 & 81.6 & 85.3 & 86.2 \\
& DOP-OBC   & \textbf{65.2}\textsuperscript{\bdelta{+4.3}} & \textbf{76.4}\textsuperscript{\bdelta{+3.3}} & \textbf{33.8}\textsuperscript{\bdelta{+1.8}} & \textbf{54.3}\textsuperscript{\bdelta{+6.2}} & \textbf{68.5}\textsuperscript{\bdelta{+3.9}} & \textbf{41.2}\textsuperscript{\bdelta{+9.0}} & \textbf{10.1}\textsuperscript{\bdelta{+5.5}} & \textbf{85.6}\textsuperscript{\bdelta{+4.0}} & \textbf{87.7}\textsuperscript{\bdelta{+2.4}} & \textbf{90.2}\textsuperscript{\bdelta{+4.0}} \\
\midrule
\multirow{2}{*}{Chat-UniVi\cite{chen2023chatunivi}}
& Base    & 56.3 & 70.4 & 28.3 & 46.9 & 59.9 & 52.3 & 16.7 & 85.1 & 69.5 & 64.4 \\
& DOP-OBC   & \textbf{63.4}\textsuperscript{\bdelta{+7.1}} & \textbf{75.9}\textsuperscript{\bdelta{+5.5}} & \textbf{33.7}\textsuperscript{\bdelta{+5.4}} & \textbf{52.4}\textsuperscript{\bdelta{+5.5}} & \textbf{64.3}\textsuperscript{\bdelta{+4.4}} & \textbf{46.2}\textsuperscript{\bdelta{+6.1}} & \textbf{14.3}\textsuperscript{\bdelta{+2.4}} & \textbf{89.3}\textsuperscript{\bdelta{+4.2}} & \textbf{70.1}\textsuperscript{\bdelta{+0.6}} & \textbf{67.3}\textsuperscript{\bdelta{+2.9}} \\
\midrule
\multirow{2}{*}{InstructBLIP \cite{dai2023instructblip}}
& Base    & 43.4 & 58.2 & 25.6 & 33.4 & 62.1 & 55.6 & 24.2 & 88.7 & 81.3 & 74.4 \\
& DOP-OBC   & \textbf{49.1}\textsuperscript{\bdelta{+5.7}} & \textbf{64.3}\textsuperscript{\bdelta{+6.1}} & \textbf{29.2}\textsuperscript{\bdelta{+3.6}} & \textbf{37.5}\textsuperscript{\bdelta{+4.1}} & \textbf{66.4}\textsuperscript{\bdelta{+4.3}} & \textbf{50.2}\textsuperscript{\bdelta{+5.4}} & \textbf{22.5}\textsuperscript{\bdelta{+1.7}} & \textbf{91.2}\textsuperscript{\bdelta{+2.5}} & \textbf{87.3}\textsuperscript{\bdelta{+6.0}} & \textbf{78.3}\textsuperscript{\bdelta{+3.9}} \\
\bottomrule
\end{tabular}}
\end{table*}

\textbf{Video benchmarks.}
For video question answering, we use MSVD-QA \cite{xu2017msvdqa} and ActivityNet-QA \cite{yu2019activitynetqa}. For open-ended video captioning, we follow recent practice and employ GPT-4o as an automatic evaluator, reporting five dimensions: Correctness (Cr.), Consistency (Cs.), Detail (De.), Context (Ct.), and Temporal understanding (Te.), with higher scores indicating better performance (see Tab.~\ref{tab:video-bench}).

\subsection{Backbone Models}
\label{sec:baselines}

\paragraph{\textbf{Image MLLMs: }}
We evaluate representative instruction-tuned image MLLMs. LLaVA-1.5 \cite{liu2024llava15} extends LLaVA with improved visual instruction tuning based on CLIP–Vicuna, but relies on standard RoPE positional encoding and can exhibit long-context drift and attention collapse, which contribute to object hallucination. InstructBLIP \cite{dai2023instructblip} combines a frozen vision encoder with a Q-Former and an instruction-tuned LLM. While broadly capable, it inherits common LLM issues such as hallucination and miscalibration, as noted by its authors \cite{dai2023instructblip}. 
% These behaviors align with FarSight’s analysis that decoder attention may over-disperse and lose positional grounding, leading to cascading “initial-to-snowball” errors \cite{tang2025farsight}.

\paragraph{\textbf{Video MLLMs: }}
For video, we consider Video-LLaVA \cite{lin2024videollava}, VILA \cite{vila2024onpretrain}, Video-LLaMA2 \cite{ma2024videollama2}, and Chat-UniVi \cite{chen2023chatunivi}. Video-LLaVA adapts LLaVA to video via frame sampling and temporal adapters but inherits similar positional and attention biases. VILA emphasizes unified vision–language pretraining improvements but does not address inference-time attention allocation. Video-LLaMA2 strengthens visual backbones and alignment for video understanding, yet relies on standard positional schemes that can drift over long temporal spans. Chat-UniVi unifies multimodal instruction following for video, but like prior MLLMs, lacks decoding-time mechanisms to balance attention across dominant and long-tail objects. As a result, visually salient content can monopolize decoding capacity while smaller or rarer objects remain underrepresented.

\paragraph{\textbf{Decoding-time baselines: }}
We additionally compare against training-free inference controls designed to improve grounding. ICD \cite{zhou2024icd} biases decoding toward image evidence, VCD \cite{wang2024vcd} integrates visual features through contrastive decoding, and OPERA \cite{li2024opera} optimizes retrieval of visual evidence during generation. While these approaches enhance faithfulness, they do not explicitly redistribute attention mass across objects, nor do they address attention collapse or positional decay effects identified by FarSight \cite{tang2025farsight}.

% \paragraph{Complementarity of our approach.}
% FarSight attributes hallucinations to two primary factors: attention mass spreading toward low-information tokens (attention collapse) and gradual positional decay over long sequences \cite{tang2025farsight}. In contrast, our DOP--OBC framework directly reshapes attention allocation during decoding by dampening over-dominant rows and boosting rare but confident ones, while reinforcing absolute positional focus through the register mechanism. As such, our method is complementary to existing decoding-time strategies and targets a distinct, equity-driven dimension of hallucination mitigation.

\subsection{Quantitative Results}
\paragraph{\textbf{Hallucination Diagnostics (CHAIR \& POPE): }}
Table~\ref{tab:positional} isolates CHAIR/ POPE under different positional schemes. Relative to RoPE and prior fixes (FixVPE/EDVT), our method achieves reductions in \textbf{CHAIRS}/\textbf{CHAIRI} and increases in \textbf{POPE-R}/\textbf{POPE-P} on both LLaVA-1.5 and Video-LLaVA. The decreases in $\text{CHAIRI}$ (instance-level) are notable, aligning with our goal of reallocating attention to small/rare objects rather than merely suppressing text length.

\paragraph{\textbf{Video QA and Open-Ended Generation: }}
Table~\ref{tab:video-bench} reports video QA accuracy/score on MSVD-QA and ActivityNet-QA, and GPT-4o-assessed text quality. Our method improves accuracy and open-ended scores for all four video MLLMs. Gains on \textbf{Cr.}/\textbf{Cs.} suggest fewer factual inconsistencies. Improvements on \textbf{Te.} indicate better temporal grounding, consistent with our per-row decay that sustains evidence across steps.

\paragraph{\textbf{Image Benchmarks: }}
Table~\ref{tab:image-bench} summarizes image-level results. Across general-ability suites (MMBench, LLaVA-W, MM-Vet, VizWiz, SQA) our method improves over the base models, while substantially reducing object-hallucination scores (CHAIR/CHAIR\textsubscript{I}) and improving POPE. For example, on LLaVA-1.5 we observe gains on hallucination diagnostics (e.g., $\downarrow$CHAIRS and $\downarrow$CHAIRI) alongside increases on POPE-R/P/A, indicating stronger grounding rather than over-conservative outputs. InstructBLIP and Video-LLaVA (image mode) show similar trends, confirming the plug-and-play nature of our approach.

\begin{figure}[!t]
\vspace{0.2cm}
    \centering
    \begin{minipage}[t]{0.48\textwidth}
        \centering
        \begin{subfigure}[t]{\textwidth}
            \centering
            \includegraphics[width=\linewidth]{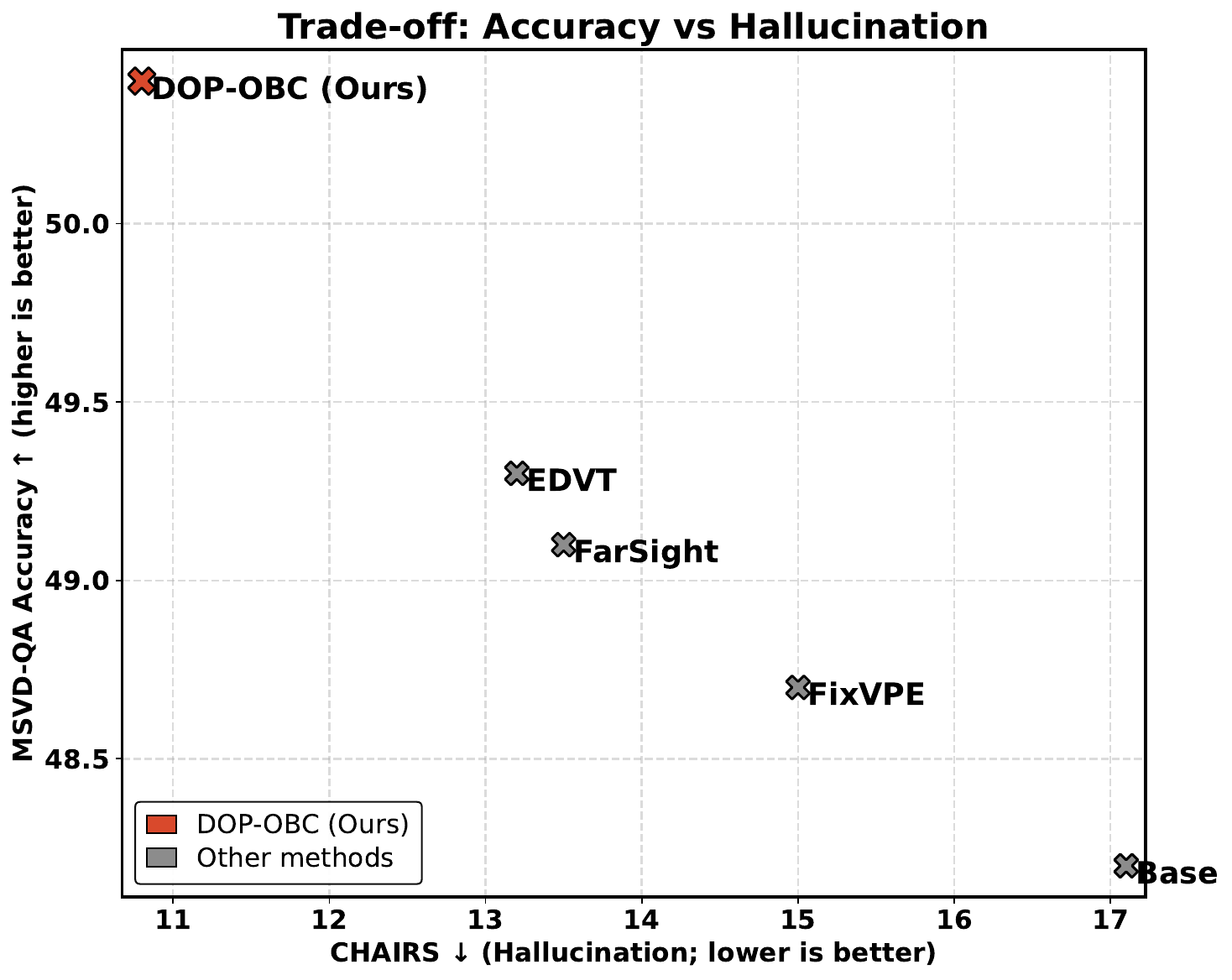}
            \caption{DOP-OBC at the Frontier}
        \end{subfigure}
        \vspace{0.3cm}
        \begin{subfigure}[t]{\textwidth}
            \centering
            \includegraphics[width=\linewidth]{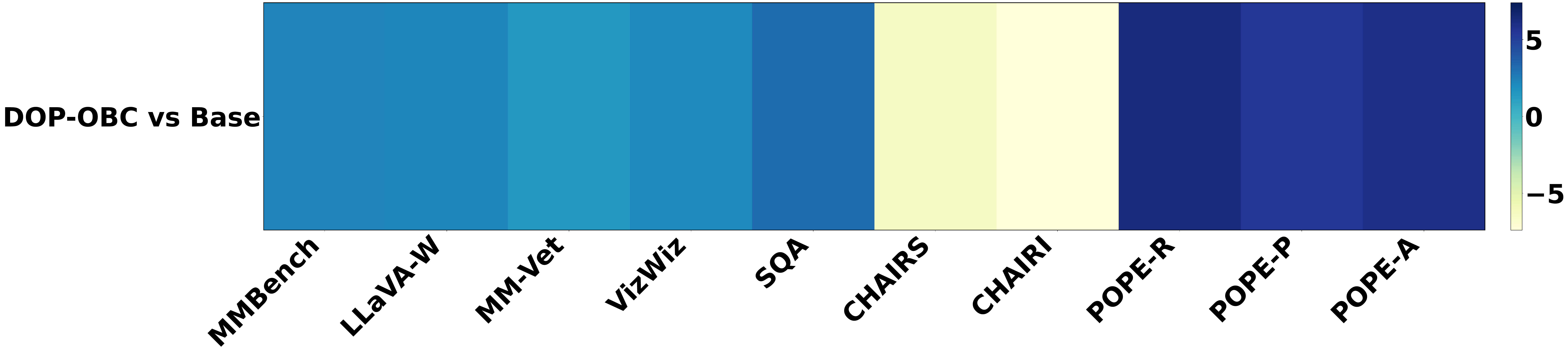}
            \caption{Relative Difference Heatmap (DOP-OBC)}
        \end{subfigure}
    \end{minipage}
    \hfill
    \begin{minipage}[t]{0.48\textwidth}
        \centering
        \begin{subfigure}[t]{\textwidth}
            \centering
            \includegraphics[width=\linewidth]{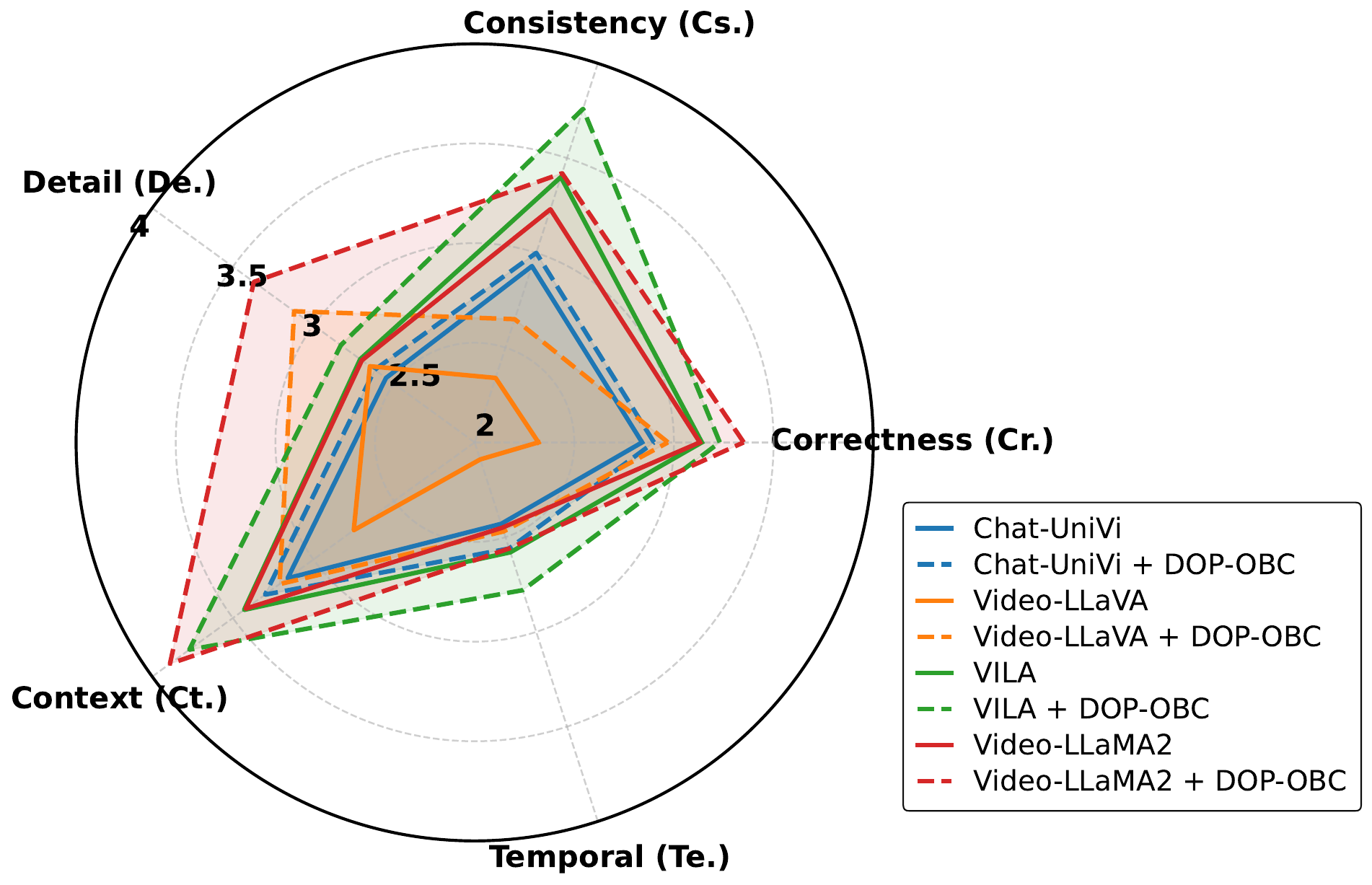}
            \caption{Equity-Aware Quality Growth}
        \end{subfigure}
    \end{minipage}

    % \caption{Detailed analysis of the impact of \textbf{DOP-OBC} on multimodal performance. (a) In the Accuracy (↑) vs. CHAIRS (↓) plane, DOP-OBC defines the empirical frontier, outperforming previous fixes. (b) Relative Difference Heatmap (DOP-OBC $|$ Base). Improvements are shown across ten benchmarks: DOP-OBC $|$ Base raises performance on general metrics and POPE scores while sharply reducing CHAIRS/CHAIRI hallucination rates. (c) Equity-Aware Quality Growth. Outward shift shows uniform gains in correctness, context, detail, and time.}
    \caption{\textbf{Comprehensive analysis of DOP--OBC’s impact on multimodal performance.}
    (a) Accuracy ($\uparrow$) vs.\ CHAIRS ($\downarrow$): DOP--OBC defines the empirical frontier, achieving higher accuracy with lower hallucination.
    (b) Relative Difference Heatmap (DOP--OBC $|$ Base): consistent gains across ten benchmarks, improving general metrics and POPE while reducing CHAIRS/CHAIRI.
    (c) Equity-Aware Quality Growth: outward shifts across Correctness, Consistency, Detail, Context, and Temporal dimensions indicate uniform improvements in video caption quality.}
    \label{fig:Figure 5}
    \vspace{-10 pt}
\end{figure}

\paragraph{\textbf{Holistic Performance Analysis:} }
While Tables~\ref{tab:image-bench}--\ref{tab:video-bench} report individual metrics, Figure~\ref{fig:Figure 5} provides a consolidated view of DOP--OBC’s impact. As shown in Figure~\ref{fig:Figure 5}(a), our method defines a new empirical frontier in the accuracy–hallucination plane, achieving higher accuracy with lower CHAIR scores compared to prior decoding fixes. Figure~\ref{fig:Figure 5}(b) illustrates consistent relative improvements across ten benchmarks, while Figure~\ref{fig:Figure 5}(c) shows uniform gains across correctness, consistency, detail, context, and temporal dimensions in video captioning. These trends indicate that equitable attention reallocation improves grounding without trading off general performance.

\begin{figure}[t]
    \centering
    \includegraphics[width=1.0\columnwidth]{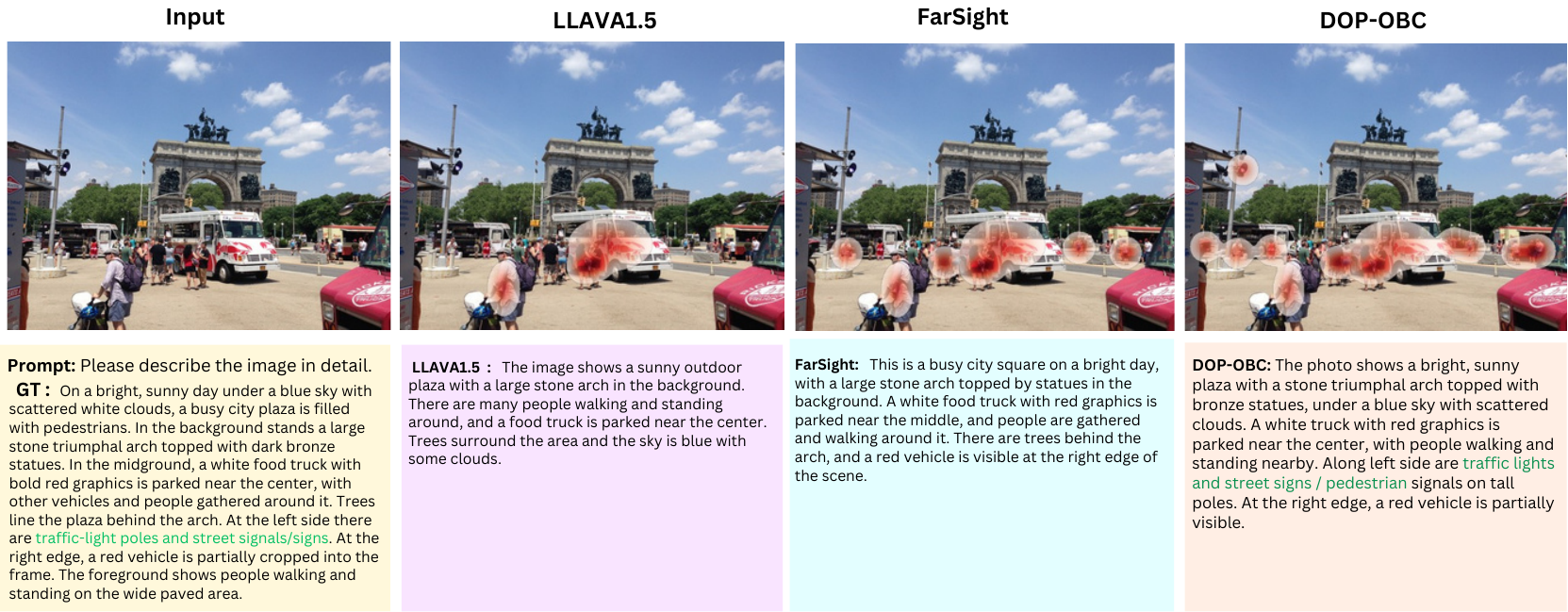}
    \vspace{-10pt}
    % \caption{\small  
    %     Qualitative comparison and attention dynamics.
    % Base model suffers attention collapse after early generation steps (right), causing it to miss peripheral objects (traffic lights, street signs, red vehicle). FarSight partially recovers but remains unstable. DOP-OBC sustains stable image attention throughout decoding, faithfully grounding all GT-mentioned objects (highlighted in red) without any model modification. 
    %     }
    \caption{\textbf{Qualitative comparison and attention dynamics during decoding.}
    \textbf{LLaVA-1.5 (base model)} exhibits early attention collapse, leading to omission of peripheral objects (e.g., traffic lights, street signs, red vehicle). FarSight partially mitigates this effect but remains unstable. \textbf{DOP--OBC} maintains stable image attention throughout generation, correctly grounding all GT-mentioned objects without architectural modification or retraining.}
    \label{fig:qualitative_fig}
    \vspace{-10 pt}
\end{figure}

\subsection{Qualitative Results}

Figure~\ref{fig:qualitative_fig} compares \textbf{LLaVA-1.5 (base)}, \textbf{FarSight} and \textbf{DOP--OBC} on a complex plaza scene containing both dominant (arch, food truck) and peripheral objects (traffic lights, street signs, red vehicle). LLaVA-1.5 captures the main structures but exhibits early attention collapse, leading to omission of peripheral yet clearly visible objects. FarSight improves object coverage but remains unstable across decoding steps, resulting in incomplete grounding.

In contrast, \textbf{DOP--OBC} maintains stable image attention throughout generation and correctly grounds all GT-mentioned objects, including peripheral elements. These qualitative results confirm that equity-aware attention reallocation mitigates attention collapse and improves completeness and visual grounding without retraining or architectural modification.

% \section{Discussion}
% \label{sec:discussion}

% Our results indicate that object hallucination in modern MLLMs is fundamentally a decoding-time attention imbalance rather than a perception deficit. Across diverse image and video backbones, DOP--OBC consistently reduces CHAIR/POPE hallucinations while improving general VQA and captioning metrics, suggesting stronger grounding rather than conservative generation. 

% The gains arise from directly reshaping decoder attention: soft suppression of over-dominant objects prevents early attention collapse, while confidence-gated amplification of rare objects preserves minority evidence across decoding steps. The positional decay mechanism further stabilizes long-range grounding without altering autoregressive causality. Importantly, these effects are achieved entirely at inference time, requiring no retraining, auxiliary supervision, or architectural modification. Overall, the findings suggest that faithful multimodal generation depends critically on how representational capacity is allocated during decoding. Equity-aware attention rebalancing offers a lightweight yet effective pathway toward improving grounding in large multimodal models.

\section{Discussion}
\label{sec:discussion}

Our results indicate that object hallucination in modern MLLMs is fundamentally a decoding-time attention imbalance rather than a perception deficit. Across diverse image and video backbones, \textbf{DOP--OBC} consistently reduces CHAIR/POPE hallucinations while improving general VQA and captioning metrics, as shown in Table~\ref{tab:image-bench}, Table~\ref{tab:video-bench}, and Figure~\ref{fig:chair-pope}. The consistent gains on both grounding and task performance support our central claim that equitable reallocation of attention improves visual grounding without degrading fluency or general reasoning.

The improvements arise from reshaping how decoder attention is distributed. By suppressing excessively attended rows, our method prevents the model from prematurely locking onto dominant content and collapsing attention early in generation. By amplifying rows associated with rare yet confident objects, it preserves minority evidence across decode steps, allowing under-represented signals to influence the output. The positional decay mechanism improves stability over long contexts by counteracting the natural flattening of attention over distance while preserving causal masking. In effect, these interventions create a more balanced attention geometry that sustains object-level evidence. The qualitative results in Figure~\ref{fig:qualitative_fig} further illustrate how DOP--OBC maintains stable and more evenly distributed attention compared to baselines.

Importantly, these benefits are achieved entirely at inference time and do not require retraining, auxiliary supervision, or architectural modification. The method is compatible with a wide range of transformer-based MLLMs, making it practical for deployed systems. While our analysis highlights robustness to the key hyperparameters and wide applicability across scenes and tasks, we also observe limitations. In very crowded scenes with dense proposals, the suppression term can become conservative and may under-represent mid-frequency objects. Noisy or low-quality object proposals can also lead to occasional amplification of spurious content, though our confidence gating mitigates this. These limitations point toward future work that more tightly integrates proposal quality and uncertainty into decode-time attention allocation.

Overall, our findings suggest that faithful multimodal generation depends critically on how representational capacity is allocated during decoding. Equity-aware attention rebalancing offers a lightweight, effective pathway to stronger grounding in large multimodal models and highlights that a significant fraction of hallucination arises from decoder attention mechanics rather than vision representation shortcomings.

\section{Conclusions}

We presented \textbf{DOP–OBC}, a lightweight, training-free decoding strategy that alleviates object hallucination in multimodal large language models by directly addressing structural imbalances in decoder attention. Instead of altering model weights, adding auxiliary supervision, or redesigning architectures, DOP–OBC rebalances attention at inference time via soft suppression of over-attended objects and calibrated amplification of rare yet confident object signals within a causal attention register.
Extensive evaluation across diverse image and video models shows that DOP–OBC consistently reduces CHAIR and POPE hallucination rates while enhancing performance on standard captioning and VQA benchmarks, indicating improved visual grounding without sacrificing descriptive quality. These results support the view that hallucination in current MLLMs arises in large part from capacity allocation during decoding. By reallocating representational resources toward under-served visual entities, equity-aware attention control offers a simple and effective means to achieve more faithful and comprehensive multimodal generation.

\noindent\textbf{Limitations.}
DOP--OBC relies on lightweight rarity and confidence signals derived from object statistics, which do not explicitly model full uncertainty or proposal reliability. In crowded scenes where objects have similar confidence, dominance cues can become less distinctive, reducing the strength of selective rebalancing. Noisy detections may also limit the precision of rarity-based amplification despite confidence gating. We focus on inference-time control and do not explore joint end-to-end optimization of the register parameters during training.

\noindent\textbf{Outlook.}
Future work can incorporate stronger uncertainty modeling and proposal calibration to further refine attention allocation. Extending equity-aware decoding to multi-image, long-video, and dialogue settings, as well as combining inference-time control with lightweight training objectives, offers promising directions for improving multimodal grounding at scale.

\clearpage  % TODO FINAL: This \clearpage needs to be removed from both review and camera-ready versions.

% \section*{Acknowledgements}
% Please insert your acknowledgments here.

% ---- Bibliography ----
%
% BibTeX users should specify bibliography style 'splncs04'.
% References will then be sorted and formatted in the correct style.
%
\bibliographystyle{splncs04}
\bibliography{main}

@String(CVPR  = {IEEE Conf. Comput. Vis. Pattern Recog.})

@String(ICCV  = {Int. Conf. Comput. Vis.})

@String(NeurIPS = {Adv. Neural Inform. Process. Syst.})

@String(CVPR  = {CVPR})

@String(ICCV  = {ICCV})

@String(NeurIPS = {NeurIPS})

@article{yu2023mm,
  title={Mm-vet: Evaluating large multimodal models for integrated capabilities},
  author={Yu, Weihao and Yang, Zhengyuan and Li, Linjie and Wang, Jianfeng and Lin, Kevin and Liu, Zicheng and Wang, Xinchao and Wang, Lijuan},
  journal={arXiv preprint arXiv:2308.02490},
  year={2023}
}

@inproceedings{rohrbach2018object,
  title={Object Hallucination in Image Captioning},
  author={Rohrbach, Anna and Hendricks, Lisa Anne and Burns, Kaylee and Darrell, Trevor and Saenko, Kate},
  booktitle={EMNLP},
  year={2018}
}

@inproceedings{li2023pope,
  title={Evaluating Object Hallucination in Large Vision-Language Models},
  author={Li, Yifan and Du, Yifan and Zhou, Kun and Wang, Jinpeng and Zhao, Wayne Xin and Wen, Ji-Rong},
  booktitle={EMNLP},
  year={2023},
  note={POPE benchmark}
}

@inproceedings{tang2025farsight,
  title={Seeing Far and Clearly: Mitigating Hallucinations in MLLMs with Attention Causal Decoding},
  author={Tang, Feilong and Liu, Chengzhi and Xu, Zhongxing and Hu, Ming and Peng, Zelin and Yang, Zhiwei and others},
  booktitle={CVPR},
  year={2025}
}

@inproceedings{liu2024llava15,
  title={Improved Baselines with Visual Instruction Tuning},
  author={Liu, Haotian and Li, Chunyuan and Li, Yuheng and Lee, Yong Jae},
  booktitle={CVPR},
  year={2024}
}

@article{liu2023llava,
  title={Visual Instruction Tuning},
  author={Liu, Haotian and Li, Chunyuan and Wu, Qingyang and Lee, Yong Jae},
  journal={arXiv:2304.08485},
  year={2023}
}

@article{dai2023instructblip,
  title={InstructBLIP: Towards General-Purpose Vision-Language Models with Instruction Tuning},
  author={Dai, Wenliang and Li, Junnan and Li, Dongxu and others},
  journal={arXiv:2305.06500},
  year={2023}
}

@inproceedings{lin2024videollava,
  title={Video-{LLaVA}: Learning United Visual Representation by Alignment Before Projection},
  author={Lin, Bin and Ye, Yang and Zhu, Bin and Cui, Jiaxi and Ning, Munan and Jin, Peng and Yuan, Li},
  booktitle={EMNLP},
  year={2024}
}

@article{zhang2023vila,
  title={{VILA}: On Pre-training for Visual Language Models},
  author={Lin, Ji and Yin, Hongxu and Ping, Wei and Lu, Yao and Molchanov, Pavlo and Tao, Andrew and Mao, Huizi and Kautz, Jan and Shoeybi, Mohammad and Han, Song},
  journal={arXiv preprint arXiv:2312.07533},
  year={2023}
}

@article{ma2024videollama2,
  title={{VideoLLaMA~2}: Advancing Spatial\textnormal{-}Temporal Modeling and Audio Understanding in Video\textnormal{-}LLMs},
  author={Cheng, Zesen and Leng, Sicong and Zhang, Hang and Xin, Yifei and Li, Xin and Chen, Guanzheng and Zhu, Yongxin and Zhang, Wenqi and Luo, Ziyang and Zhao, Deli and Bing, Lidong},
  journal={arXiv preprint arXiv:2406.07476},
  year={2024}
}

@article{chen2023chatunivi,
  title={Chat\textnormal{-}UniVi: Unified Visual Representation Empowers Large Language Models with Image and Video Understanding},
  author={Jin, Peng and Takanobu, Ryuichi and Zhang, Caiwan and Cao, Xiaochun and Yuan, Li},
  journal={arXiv preprint arXiv:2311.08046},
  year={2023}
}

@article{liu2023mmbench,
  title={{MMBench}: Is Your Multi-Modal Language Model Getting Better?},
  author={Liu, Haotian and others},
  journal={arXiv preprint arXiv:2307.06281},
  year={2023}
}

@inproceedings{gurari2018vizwiz,
  title={VizWiz Grand Challenge: Answering Visual Questions from Blind People},
  author={Gurari, Danna and others},
  booktitle={CVPR},
  year={2018}
}

@inproceedings{lu2022scienceqa,
  title={Learn to Explain: Multimodal Reasoning via Thought Chains for Science {QA}},
  author={Lu, Pan and others},
  booktitle={NeurIPS},
  year={2022}
}

@inproceedings{xu2017msvdqa,
  title={{MSVD-QA}: A Large-Scale Video {QA} Dataset for Open-Ended Understanding},
  author={Xu, Jun and others},
  booktitle={ICCV},
  year={2017}
}

@inproceedings{yu2019activitynetqa,
  title={ActivityNet-{QA}: A Dataset for Video Question Answering},
  author={Yu, Zhou and others},
  booktitle={CVPR},
  year={2019}
}

@InProceedings{wang2024vcd,
  author    = {Leng, Sicong and Zhang, Hang and Chen, Guanzheng and Li, Xin and Lu, Shijian and Miao, Chunyan and Bing, Lidong},
  title     = {Mitigating Object Hallucinations in Large Vision-Language Models Through Visual Contrastive Decoding},
  booktitle = {Proceedings of the IEEE/CVF Conference on Computer Vision and Pattern Recognition (CVPR)},
  month     = {June},
  year      = {2024},
  pages     = {13872--13882}
}

@inproceedings{leng2024mitigating,
  title={Mitigating object hallucinations in large vision-language models through visual contrastive decoding},
  author={Leng, Sicong and Zhang, Hang and Chen, Guanzheng and Li, Xin and Lu, Shijian and Miao, Chunyan and Bing, Lidong},
  booktitle={Proceedings of the IEEE/CVF Conference on Computer Vision and Pattern Recognition},
  pages={13872--13882},
  year={2024}
}

@article{zhou2024icd,
  title   = {Mitigating Hallucinations in Large Vision-Language Models with Instruction Contrastive Decoding},
  author  = {Wang, Xintong and Pan, Jingheng and Ding, Liang and Biemann, Chris},
  journal = {arXiv preprint arXiv:2403.18715},
  year    = {2024}
}

@article{alayrac2022flamingo,
  title={Flamingo: a visual language model for few-shot learning},
  author={Alayrac, Jean-Baptiste and Donahue, Jeff and Luc, Pauline and Miech, Antoine and Barr, Iain and Hasson, Yana and Lenc, Karel and Mensch, Arthur and Millican, Katherine and Reynolds, Malcolm and others},
  journal={Advances in neural information processing systems},
  volume={35},
  pages={23716--23736},
  year={2022}
}

@InProceedings{li2024opera,
  author    = {Huang, Qidong and Dong, Xiaoyi and Zhang, Pan and Wang, Bin and He, Conghui and Wang, Jiaqi and Lin, Dahua and Zhang, Weiming and Yu, Nenghai},
  title     = {OPERA: Alleviating Hallucination in Multi-Modal Large Language Models via Over-Trust Penalty and Retrospection-Allocation},
  booktitle = {Proceedings of the IEEE/CVF Conference on Computer Vision and Pattern Recognition (CVPR)},
  month     = {June},
  year      = {2024},
  pages     = {13418--13427}
}

@article{vila2024onpretrain,
  title={{VILA}: On Pre-training for Visual Language Models},
  author={Lin, Ji and Yin, Hongxu and Ping, Wei and Lu, Yao and Molchanov, Pavlo and Tao, Andrew and Mao, Huizi and Kautz, Jan and Shoeybi, Mohammad and Han, Song},
  journal={arXiv preprint arXiv:2312.07533},
  year={2023}
}

@inproceedings{yin2025clearsight,
  title={Clearsight: Visual signal enhancement for object hallucination mitigation in multimodal large language models},
  author={Yin, Hao and Si, Guangzong and Wang, Zilei},
  booktitle={Proceedings of the Computer Vision and Pattern Recognition Conference},
  pages={14625--14634},
  year={2025}
}

@article{wang2025ascd,
  title={ASCD: Attention-Steerable Contrastive Decoding for Reducing Hallucination in MLLM},
  author={Wang, Yujun and Bi, Jinhe and Pirk, Soeren and Ma, Yunpu and others},
  journal={arXiv preprint arXiv:2506.14766},
  year={2025}
}

@inproceedings{liu2025multi,
  title={Multi-Frequency Contrastive Decoding: Alleviating Hallucinations for Large Vision-Language Models},
  author={Liu, Bingqian and Zhang, Fu and Chen, Guoqing and Cheng, Jingwei},
  booktitle={Proceedings of the 2025 Conference on Empirical Methods in Natural Language Processing},
  pages={28556--28572},
  year={2025}
}

@inproceedings{xu2025mitigating,
  title={Mitigating hallucinations in multi-modal large language models via image token attention-guided decoding},
  author={Xu, Xinhao and Chen, Hui and Lyu, Mengyao and Zhao, Sicheng and Xiong, Yizhe and Lin, Zijia and Han, Jungong and Ding, Guiguang},
  booktitle={Proceedings of the 2025 Conference of the Nations of the Americas Chapter of the Association for Computational Linguistics: Human Language Technologies (Volume 1: Long Papers)},
  pages={1571--1590},
  year={2025}
}

@inproceedings{wang2025scaling,
  title={Scaling inference-time search with vision value model for improved visual comprehension},
  author={Wang, Xiyao and Yang, Zhengyuan and Li, Linjie and Lu, Hongjin and Xu, Yuancheng and Lin, Chung-Ching and Lin, Kevin and Huang, Furong and Wang, Lijuan},
  booktitle={Proceedings of the IEEE/CVF International Conference on Computer Vision},
  pages={1173--1184},
  year={2025}
}

@article{deria2025dual,
  title={Dual-Stage Value-Guided Inference with Margin-Based Reward Adjustment for Fast and Faithful VLM Captioning},
  author={Deria, Ankan and Dukre, Adinath Madhavrao and Tang, Feilong and Atito, Sara and Roy, Sudipta and Awais, Muhammad and Khan, Muhammad Haris and Razzak, Imran},
  journal={arXiv preprint arXiv:2506.15649},
  year={2025}
}

@inproceedings{dong2025inter,
  title={INTER: Mitigating Hallucination in Large Vision-Language Models by Interaction Guidance Sampling},
  author={Dong, Xin and Dong, Shichao and Wang, Jin and Huang, Jing and Zhou, Li and Sun, Zenghui and Jing, Lihua and Lan, Jinsong and Zhu, Xiaoyong and Zheng, Bo},
  booktitle={Proceedings of the IEEE/CVF International Conference on Computer Vision},
  pages={2534--2544},
  year={2025}
}

@article{deng2024seeing,
  title={Seeing is believing: Mitigating hallucination in large vision-language models via clip-guided decoding},
  author={Deng, Ailin and Chen, Zhirui and Hooi, Bryan},
  journal={arXiv preprint arXiv:2402.15300},
  year={2024}
}

@inproceedings{min2025mitigating,
  title={Mitigating hallucinations in large vision-language models via summary-guided decoding},
  author={Min, Kyungmin and Kim, Minbeom and Lee, Kang-il and Lee, Dongryeol and Jung, Kyomin},
  booktitle={Findings of the Association for Computational Linguistics: NAACL 2025},
  pages={4183--4198},
  year={2025}
}
\end{document}